\documentclass[english,american,letterpaper, 10 pt, journal, twoside]{IEEEtran}
\usepackage[T1]{fontenc}
\usepackage[utf8]{inputenc}
\usepackage[letterpaper]{geometry}
\usepackage{color}
\usepackage{babel}
\usepackage{array}
\usepackage{multirow}
\usepackage{amsmath}
\usepackage{amssymb}
\usepackage{graphicx}
\usepackage[unicode=true,hyperfootnotes=false]{hyperref}

\makeatletter

\providecommand{\tabularnewline}{\\}

\newcommand{\addrevision}[1]{#1}
\newcommand{\addrevisiongraphics}[1]{\fcolorbox{white}{white}{#1}}

\newcommand{\addlastrevision}[1]{#1}
\newcommand{\addlastRevision}[1]{#1}
\usepackage{array}

\usepackage{color}

\usepackage{upquote}

\usepackage[usenames,dvipsnames]{xcolor}
\usepackage{scalerel}
\usepackage{tikz}
\usetikzlibrary{svg.path}

\definecolor{orcidlogocol}{HTML}{A6CE39}
\tikzset{
    orcidlogo/.pic={
        \fill[orcidlogocol] svg{M256,128c0,70.7-57.3,128-128,128C57.3,256,0,198.7,0,128C0,57.3,57.3,0,128,0C198.7,0,256,57.3,256,128z};
        \fill[white] svg{M86.3,186.2H70.9V79.1h15.4v48.4V186.2z}
        svg{M108.9,79.1h41.6c39.6,0,57,28.3,57,53.6c0,27.5-21.5,53.6-56.8,53.6h-41.8V79.1z M124.3,172.4h24.5c34.9,0,42.9-26.5,42.9-39.7c0-21.5-13.7-39.7-43.7-39.7h-23.7V172.4z}
        svg{M88.7,56.8c0,5.5-4.5,10.1-10.1,10.1c-5.6,0-10.1-4.6-10.1-10.1c0-5.6,4.5-10.1,10.1-10.1C84.2,46.7,88.7,51.3,88.7,56.8z};
    }
}

\newcommand\orcidicon[1]{\href{https://orcid.org/#1}{\mbox{\scalerel*{
                \begin{tikzpicture}[yscale=-1,transform shape]
                \pic{orcidlogo};
                \end{tikzpicture}
            }{|}}}}

\usepackage[T1]{fontenc}

\usepackage{listings}
\usepackage[most]{tcolorbox}

\usepackage{caption}
\usepackage{graphics}
\usepackage{placeins}
\usepackage{graphicx, epstopdf}
\usepackage[framed, numbered]{matlab-prettifier}
\usepackage{colortbl}
\usepackage{empheq}
\usepackage{footmisc} %

\definecolor{celadon}{rgb}{0.67, 0.88, 0.69}
\definecolor{hellgelb}{rgb}{1,1,0.85} 
\definecolor{colKeys}{rgb}{0,0,1} 
\definecolor{colIdentifier}{rgb}{0,0,0} 
\definecolor{colComments}{rgb}{0,0.5,0} 
\definecolor{colString}{rgb}{0.81,0.12,0.95}

\definecolor{deepblue}{rgb}{0,0,1}
\definecolor{deepred}{rgb}{0.6,0,0}
\definecolor{deepgreen}{rgb}{0,0.5,0}
\definecolor{blue_light}{RGB}{16,161,239}

\hypersetup{
  colorlinks,
  citecolor=black,
  linkcolor=black,
  urlcolor=black}

\usepackage[algo2e,ruled,vlined,linesnumbered]{algorithm2e}
\usepackage{subfig}
\usepackage{caption}

\usepackage[dvipsnames]{xcolor}

\usepackage{boldline} 

\usepackage{diagbox}

\usepackage{colortbl}
\usepackage{tikz}
\usepackage{tabularx}
\usepackage{dcolumn, array, booktabs}

\usepackage{siunitx}

\usepackage[noadjust]{cite} %

\@ifundefined{showcaptionsetup}{}{%
 \PassOptionsToPackage{caption=false}{subfig}}
\usepackage{subfig}
\makeatother

\IEEEaftertitletext{\vspace{-1.8\baselineskip}}
\usepackage[compact]{titlesec}
\titlespacing{\section}{0pt}{0.7ex}{0.5ex}
\titlespacing{\subsection}{0pt}{0.6ex}{0ex}
\titlespacing{\subsubsection}{0pt}{0.5ex}{0ex}

\begin{document}
\title{Deep-PANTHER: Learning-Based Perception-Aware Trajectory Planner in Dynamic Environments}
\author{Jesus Tordesillas\textsuperscript{\orcidicon{0000-0001-6848-4070}} and Jonathan P. How\textsuperscript{\orcidicon{0000-0001-8576-1930}}    \thanks{Manuscript received: July 6, 2022; Revised: November 30, 2022; Accepted: January 1, 2023}%
		\thanks{
		This paper was recommended for publication by
		Bera, Aniket upon evaluation of the Associate Editor and Reviewers’ comments. %
		Research supported in part by Boeing Research \& Technology, the Air Force Office of Scientific Research MURI FA9550-19-1-0386, and the STTR with SSCI on Intelligent, Fast Reinforcement Learning for ISR Tasking (IFRIT).}
	\thanks{The authors are with the Aerospace Controls Laboratory, MIT, 77 Massachusetts
Ave., Cambridge, MA, USA \tt{\{jtorde, jhow}\}@mit.edu}	\thanks{Digital Object Identifier (DOI): see top of this page.}\vspace{-0.2cm}}

\maketitle
\renewcommand{\lstlistingname}{Script}

\newcommand{\tikzcircle}[2][black,fill=red]{\tikz[baseline=0.0ex, line width=0.3mm]\draw[#1] [#1] (0,0.08) circle (0.09);}%
\newcommand{\tikzrectangleDeepPanther}[2][black,fill=red]{\tikz[baseline=0.0ex, line width=0.2mm]\draw[#1] [#2] (0,0) rectangle (0.2,0.2);}%

\newcommand{\jhmargin}[2]{{\color{orange}#1}\marginpar{\color{orange}\tiny\raggedright \bf [JH] #2}}

\hyphenation{decentralized generate B-Spline challenging}

\markboth{IEEE Robotics and Automation Letters. Preprint Version. Accepted January, 2023}
{Tordesillas \MakeLowercase{\textit{et al.}}: Deep-PANTHER: Learning-Based Perception-Aware Trajectory Planner}

\begin{abstract}
	This paper presents Deep-PANTHER, a learning-based perception-aware trajectory planner for unmanned aerial vehicles (UAVs) in dynamic environments. 
	Given the current state of the UAV, and the predicted trajectory and size of the obstacle, Deep-PANTHER generates multiple 
	trajectories to avoid a dynamic obstacle while simultaneously maximizing its presence in the field of view~(FOV) of the onboard camera. 
	To obtain a computationally tractable real-time solution, imitation learning is leveraged to train a Deep-PANTHER policy using demonstrations provided by a multimodal optimization-based expert.
	Extensive simulations show replanning times that are two orders of magnitude faster than the optimization-based expert,
	while achieving a similar cost. 
	By ensuring that each expert trajectory is assigned to one distinct student trajectory in the loss function, Deep-PANTHER can also capture the multimodality of the problem and achieve a mean squared error (MSE) loss with respect to the expert that is up to 18 times smaller than state-of-the-art  (Relaxed)~Winner-Takes-All approaches. 
	Deep-PANTHER is also shown to generalize well to obstacle trajectories that differ from the ones used in training. 
\end{abstract}

\vspace{-0.4cm}
\begin{IEEEkeywords} UAV, Imitation Learning, Perception-Aware Trajectory Planning, Optimization. \end{IEEEkeywords}
\IEEEpeerreviewmaketitle

\noindent
\textbf{Video}: \url{http://youtu.be/53GBjP1jFW8} \\
\textbf{Code}: \url{https://github.com/mit-acl/deep_panther}

\section{Introduction and Related Work}\label{sec:introduction_deep_panther}

\definecolor{color_formulation}{RGB}{179,227,204}
\definecolor{color_goal}{RGB}{245,247,167}

\newcommand{\velunits}{m/s}
\newcommand{\accelunits}{m/s\textsuperscript{2}}
\newcommand{\jerkunits}{m/s\textsuperscript{3}}

\newcommand{\circleform}[1]{\circled{A.#1}{color_formulation!50}{black}{black}{0.6}}%
\newcommand{\circlegoal}[1]{\circled{B.#1}{color_goal!50}{black}{black}{0.6}}%

\newcommand{\poscpts}{\ensuremath{\mathcal{Q}_{\mathbf{p}}}}
\newcommand{\poscptsObst}{\ensuremath{\mathcal{Q}_{\mathbf{p},\text{obst}}}}
\newcommand{\sizeObst}{\ensuremath{\boldsymbol{s}_{\text{obst}}}}
\newcommand{\yawcpts}{\ensuremath{\mathcal{Q}_{\psi}}}
\newcommand{\poscptscropped}{\ensuremath{\widehat{\mathcal{Q}}_{\mathbf{p}}}}
\newcommand{\yawcptscropped}{\ensuremath{\widehat{\mathcal{Q}}_{\psi}}}
\newcommand{\pantherStar}{PANTHER\textsuperscript{*}}
\newcommand{\posObstacleWorldFrame}{\ensuremath{\mathbf{p}_{\text{obst}}}}
\newcommand{\posBodyWorldFrame}{\ensuremath{\mathbf{p}}}

\newcommand{\citenoPA}{\cite{chen2016online,gao2020teach,tordesillas2019faster,lin2020robust,falanga2020dynamic,wang2021autonomous,sanket2020evdodgenet}}
\newcommand{\citePAHw}{\cite{nageli2017real,bonatti2018autonomous,chen2020bio,ding2019efficient}}
\newcommand{\citePADec}{\cite{zhou2020raptor,spasojevic2020perception,murali2019perception}}
\newcommand{\citePAJoint}{\cite{watterson2020trajectory,falanga2018pampc,penin2018vision}}
\newcommand{\citeReduceStateEst}{\cite{spasojevic2020perception,murali2019perception,falanga2018pampc,watterson2020trajectory,spasojevic2020joint,bartolomei2020perception,lee2020aggressive,zhang2018perception,costante2016perception,achtelik2014motion,penin2017vision,preiss2018simultaneous,frey2019towards,salaris2019online}}
\newcommand{\citeRecordTarget}{\cite{thomas2017autonomous,jeon2020detection,penin2018vision,guanrui2021pcmpc,penin2017vision,chen2017using,jeon2019online}}

\IEEEPARstart{T}{rajectory} planning for UAVs in unknown dynamic environments is extremely challenging due to the need for gaining information about the obstacles while avoiding them at the same time. Perception-aware planning has emerged as one promising approach for this, where the translation and/or rotation of the UAV are optimized to maximize the 
presence of the obstacles in the FOV of the onboard camera 
while flying towards the goal~\cite{zhou2020raptor,spasojevic2020perception,murali2019perception,watterson2020trajectory,falanga2018pampc,penin2018vision,tordesillas2021panther}. The  dynamic nature of these environments requires very fast replanning times, which are usually achieved by simplifying the optimization problem by fixing some variables (such as the time allocation or the planes that separate the UAV from the obstacles) beforehand or by ignoring the multimodality of the problem~\cite{tordesillas2021panther}. While these simplifications help reduce the computation time, that is often achieved at the expense of more conservative planned trajectories. This leaves open the question of whether or not it is possible to obtain \textit{faster} computation times while achieving \textit{less conservative} trajectories.

\begin{figure}
	\definecolor{lightblue}{RGB}{0,215,255}
	\begin{centering}
		\includegraphics[width=1\columnwidth]{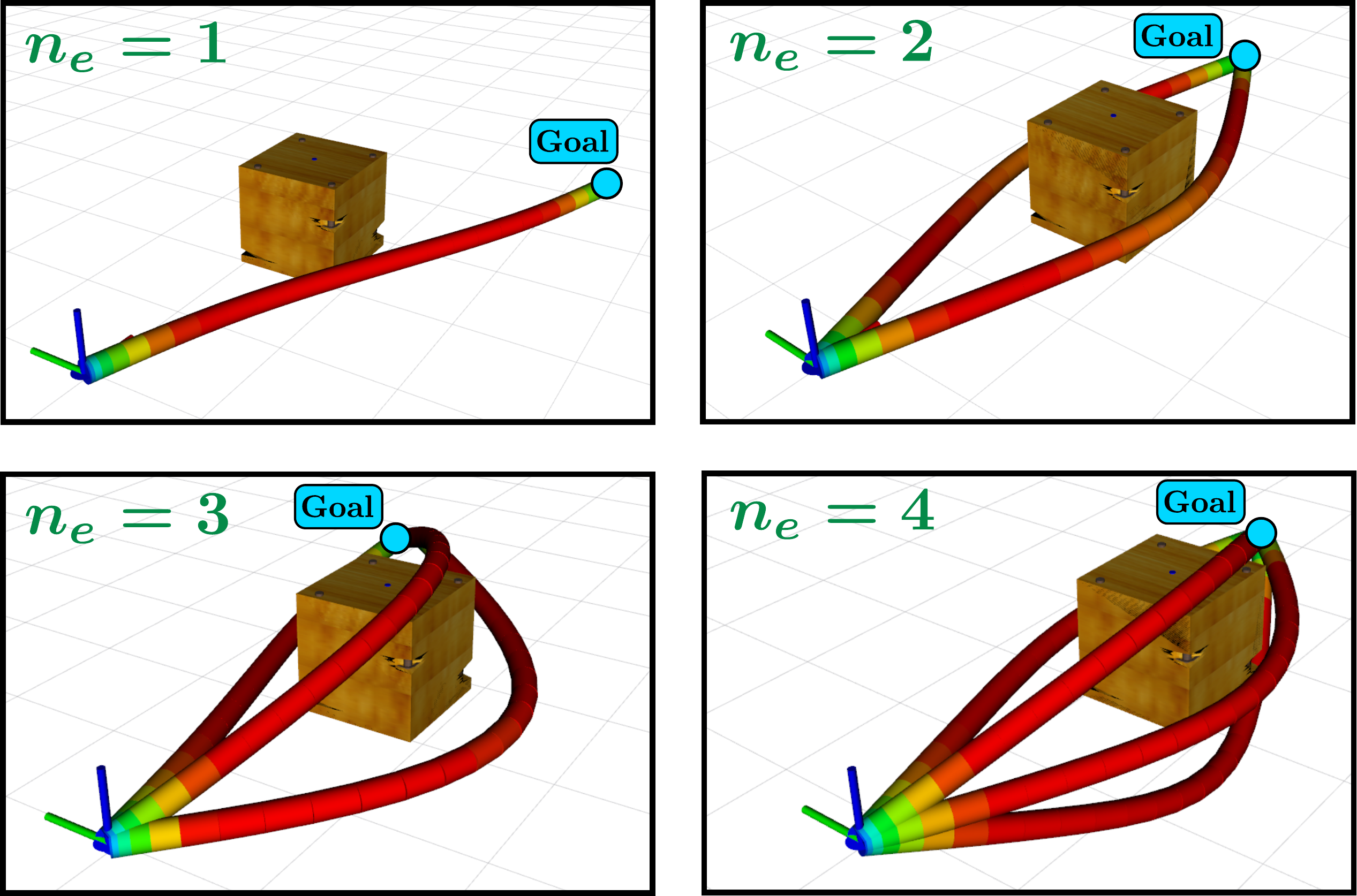}
		\par\end{centering}
	\caption[Trajectories generated by the expert for four different goals]{Trajectories generated by the expert for four different goals \tikzcircle[black,fill=lightblue]{25pt}.
		For each of the cases, we plot all the distinct locally optimal solutions found by running a total of 10 optimizations with different initial guesses. 
		The colormap represents the velocity (red denotes a higher velocity). 
		\label{fig:modes_expert}}
	\vskip-3.8ex
\end{figure}

\begin{figure*}
	\newcommand{\BlackHLine}[2][]{\tikz[baseline=0.0ex]\draw [black,thick] (0,0.08) -- (0.5,0.08);}
	\definecolor{myexpert1}{RGB}{34, 152, 94}
	\definecolor{myexpert2}{RGB}{78, 221, 95}
	\definecolor{myexpert3}{RGB}{188, 247, 218}
	
	\definecolor{mystudent1}{RGB}{255, 0, 0}
	\definecolor{mystudent2}{RGB}{255, 119, 119}
	\definecolor{mystudent3}{RGB}{255, 185, 185}
	\begin{centering}
		\includegraphics[width=1\textwidth]{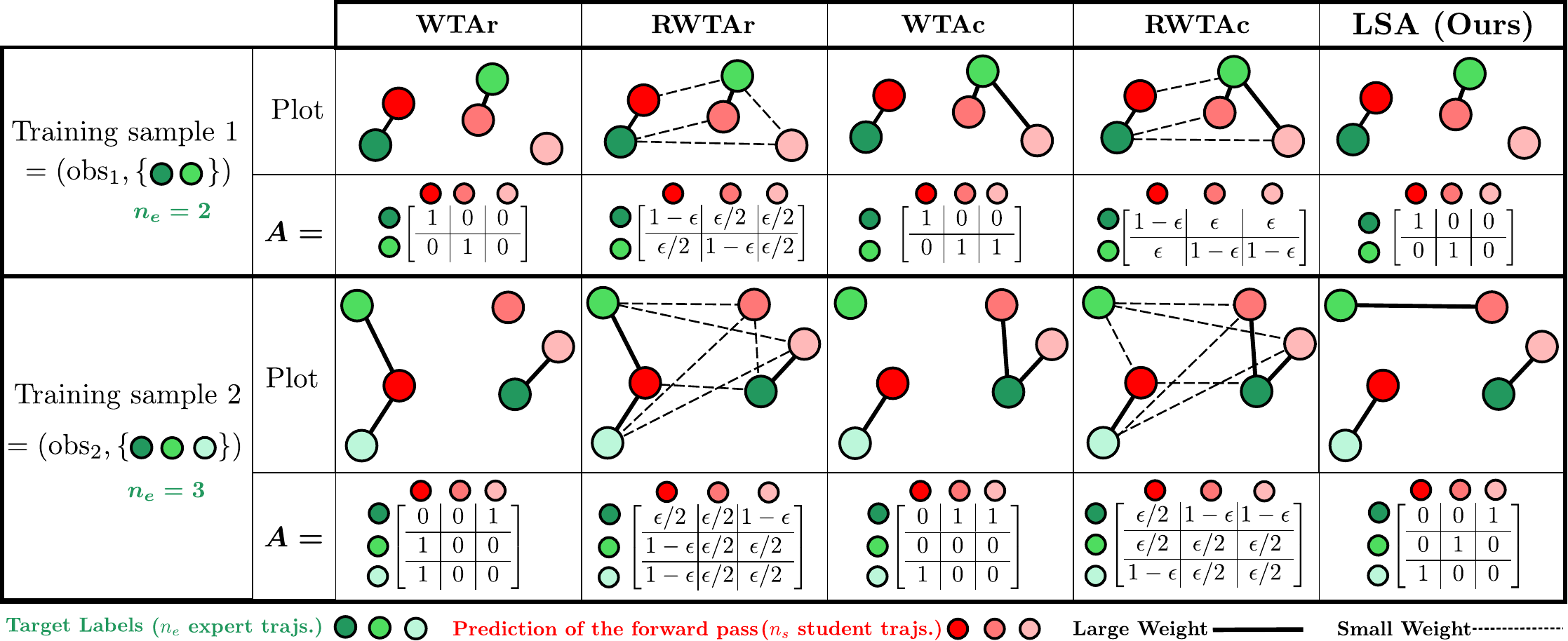}
		\par\end{centering}
	\caption{Comparison between the assignment matrix $\boldsymbol{A}$ obtained by the \addrevision{WTAr~\cite{firman2018diversenet, makansi2019overcoming}, RWTAr~\cite{makansi2019overcoming, rupprecht2017learning}, WTAc~\cite{loquercio2021learning}, RWTAc~\cite{loquercio2021learning},} and LSA approaches. This matrix $\boldsymbol{A}$ is then the one used to weigh each (target, prediction) pair in the loss. 
	In the figure, $\epsilon\ge0$, $n_s=3$ and $\text{obs}_i$ denotes the observation of the training sample $i$. In WTA\textbf{r} and RWTA\textbf{r}, each \textbf{r}ow of $\boldsymbol{A}$ sums up to~1, while in WTA\textbf{c} and RWTA\textbf{c}, each \textbf{c}olumn of $\boldsymbol{A}$ sums up to~1. We propose instead to obtain $\boldsymbol{A}$ as the solution of the linear sum assignment (LSA) problem, which minimizes the total assignment cost, and guarantees that all the target labels have one distinct prediction assigned to them (i.e., all the rows sum up to~1, $n_e$ columns sum up to~1, and $(n_s-n_e)$ columns sum up to~0). More visualizations of the WTAr and RWTAr \addrevision{assignments are available 
	at~\cite{makansi2019overcoming} %
	and~\cite{firman2018diversenet}}%
	.\label{fig:wta_rwta_ours_explanation}}
		\vskip-3.7ex
\end{figure*}

Towards this end, Imitation Learning (IL) has recently gained interest due to its ability to train a computationally-cheap neural network (the student) to approximate the solution of a computationally-expensive algorithm (the expert). IL has been successfully used to compress MPC policies~\cite{tagliabue2021efficient,kaufmann2020deep, reske2021imitation, pan2020imitation} and/or to learn path planning policies~\cite{ross2013learning, loquercio2021learning, codevilla2018end}. 
Compared to other IL-based trajectory planning works, which typically either assume static worlds or do not take into account perception awareness, our work proposes to use IL to obtain \textit{perception-aware} trajectories that perform \textit{obstacle avoidance} in \textit{dynamic} environments. 

When performing obstacle avoidance, capturing the multimodality of the trajectory planning problem is crucial to reduce the conservativeness.
Indeed, for a given scenario, there may be $n_e\ge1$ locally-optimal expert trajectories that avoid the obstacle(s)
(e.g., see Fig.~\ref{fig:modes_expert}), where $n_e$ may change between different scenarios. 
The use of a unimodal student that produces a single trajectory either introduces an artificial bias towards a specific direction of the space,
or averages together the different expert trajectories, which can be catastrophic in obstacle avoidance scenarios.
The challenge is then how to design and train a neural network capable of generating a multimodal trajectory prediction.%

One possible approach is to use  Mixture Density Networks to learn the parameters of a Gaussian mixture model~\cite{bishop1994mixture}. Mixture Density networks are however known to suffer from numerical instability and mode collapse~\cite{rupprecht2017learning, makansi2019overcoming}. 
\addrevision{Another option is to design multimodal losses\footnote{\addrevision{In this paper we use the term \textit{multimodal} to refer to the fact that the set of predicted trajectories can contain more than one trajectory. Intuitively, this means that the planned trajectories capture the fact that we can \textit{go right, left, up,...} (see Fig.~\ref{fig:modes_expert}). Unimodal approaches, on the contrary, are able to generate only one trajectory. }} that are able to compare a set of predicted trajectories with a set of target trajectories. For example,} the Winner-Takes-All (WTAr or WTAc) losses~\cite{guzman2012multiple, firman2018diversenet, makansi2019overcoming} (see Fig.~\ref{fig:wta_rwta_ours_explanation}) %
use an binary assignment matrix $\boldsymbol{A}$ that weighs the contribution of each (target, prediction) pair in the loss. 
In WTAr~\cite{firman2018diversenet, makansi2019overcoming} \addrevision{ (Winner-Takes-All-row)}, each target label is assigned to the closest prediction, while in WTAc \addrevision{ (Winner-Takes-All-column)}, each prediction is assigned to the closest target label. Other works propose instead the use of the relaxed losses RWTAr~\cite{makansi2019overcoming, rupprecht2017learning} \addrevision{ (Relaxed-Winner-Takes-All-row)} and RWTAc~\cite{loquercio2021learning} \addrevision{ (Relaxed-Winner-Takes-All-column)}, where the constraint of $\boldsymbol{A}$ being a binary matrix is relaxed (see Fig.~\ref{fig:wta_rwta_ours_explanation}). 
These relaxed costs typically address the mode collapse problem (which happens when all the predictions of the network after training are close to the same target label), but due to the nonzero weights between all the predictions and all the target labels, the predictions of these relaxed costs may reach an equilibrium position that does not represent any of the target labels~\cite{narayanan2021divide}.

In contrast to these approaches, and inspired by the multi-object detection and tracking algorithms~\cite{xu2020train, bewley2016simple, carion2020end}, we propose to use (in the loss) the optimal assignment matrix~$\boldsymbol{A}$ found by solving the linear sum assignment (LSA) problem, which minimizes the total assignment cost and guarantees that \textit{all} the target labels are assigned to a \textit{distinct} prediction (see Fig.~\ref{fig:wta_rwta_ours_explanation}).  
This ensures that a target label is not assigned to multiple predictions
(reducing therefore the mode collapse problem) and that each prediction is not assigned to multiple target labels (being therefore less prone to equilibrium issues).

The contributions of this work are therefore summarized as follows:

\begin{itemize}
	\item Novel multimodal learning-based trajectory planning framework able to generate collision-free trajectories that avoid a dynamic obstacle while maximizing its presence in the FOV. 
	\item Computation times two orders of magnitude faster than a multimodal optimization-based planner, while achieving a similar total cost \addrevision{(as defined in Section~\ref{sec:testing})}.%
	\item Multimodal loss that achieves an MSE of the predicted trajectories with respect to the expert trajectories up to $18$ times smaller than the (Relaxed)~Winner-Takes-All approaches.
	\item Deep-PANTHER also presents a very good generalization to environments where the obstacle is following a different trajectory than the one used in training.
\end{itemize}

\newcommand{\NextTrajDeepPanther}{\tikz[baseline=0.0ex]\draw [red,thick, dash pattern=on 3pt off 1pt] (0,0.08) -- (0.5,0.08);}
\newcommand{\CurrTrajDeepPanther}{\tikz[baseline=0.0ex]\draw [red,thick] (0,0.08) -- (0.5,0.08);}
\begin{table}
	\begin{centering}
		\caption[Notation used in this chapter]{Notation used in this paper.\label{tab:Notation_deep_panther}}
		\par\end{centering}
	\noindent\resizebox{1.0\columnwidth}{!}{%
		\begin{centering}
			\begin{tabular}{|>{\centering}m{0.25\columnwidth}|>{\raggedright}m{0.96\columnwidth}|}
				\hline 
				\textbf{Symbol} & \textbf{\qquad \qquad \qquad \qquad \qquad \qquad Meaning}\tabularnewline
				\hline 
				\hline 
				$\mathcal{S}_{p,m}^{d}$  & Set of clamped uniform splines with dimension $d$, degree $p$, and
				$m+1$ knots.\tabularnewline				
				\hline 
				$\circ$&Quaternion multiplication. \tabularnewline
				\hline
				$\odot$ & Element-wise product. \tabularnewline
				\hline
				$\mathcal{U}(a,b)$ & Uniform distribution in $[a,b]$.\tabularnewline
				\hline 
				$(\boldsymbol{a})_n$, $\bar{\boldsymbol{a}}$ & Vector $\boldsymbol{a}$ normalized: $(\boldsymbol{a})_n\equiv \bar{\boldsymbol{a}}:=\frac{\boldsymbol{a}}{\left\Vert \boldsymbol{a} \right\Vert}$.\tabularnewline
				\hline 
				$\boldsymbol{p}^{a}$ & Point expressed in the frame $a$. For the definitions of this table that include the sentence \emph{``expressed in the world frame''}, the notation of the frame is omitted.\tabularnewline
				\hline 				 
				$\arraycolsep=1.4pt\boldsymbol{T}_{b}^{a}=\left[\begin{array}{cc}
					\boldsymbol{R}_{b}^{a} & \boldsymbol{t}_{b}^{a}\\
					\boldsymbol{0}^{T} & 1
				\end{array}\right]$ & Transformation matrix: $\arraycolsep=1.4pt\left[\begin{array}{c}
					\boldsymbol{p}^a\\
					1
				\end{array}\right]=\boldsymbol{T}_{b}^{a}\left[\begin{array}{c}
					\boldsymbol{p}^b\\
					1
				\end{array}\right]$.
				Analogous definition for the quaternion~$\boldsymbol{q}_{b}^{a}$.\tabularnewline
				\hline 
				$\boldsymbol{e}_{x}$,
				$\boldsymbol{e}_{z}$, $\boldsymbol{1}$&$\arraycolsep=1.4pt\boldsymbol{e}_{x}:=\left[\begin{array}{ccc}1 & 0 & 0\end{array}\right]^{T}$, $\arraycolsep=1.4pt\boldsymbol{e}_{z}:=\left[\begin{array}{ccc}0 & 0 & 1\end{array}\right]^{T}$, $\arraycolsep=1.4pt\boldsymbol{1}:=\left[\begin{array}{cccc}1 & 1 & \hdots & 1\end{array}\right]^{T}$. \tabularnewline
				\hline 
				FOV, MSE, LSA&Field of View, Mean Squared Error, linear sum assignment. \tabularnewline
				
				\hline 
				\posBodyWorldFrame{}  & Position of the body frame expressed in the world frame. I.e., $\posBodyWorldFrame{}:=\boldsymbol{t}^w_b$. \tabularnewline
				\hline
				$\mathbf{a}$  & Acceleration of the body frame w.r.t. the world frame, and expressed in the world frame. \tabularnewline
				\hline 
				\posObstacleWorldFrame{} & Mean of the predicted position of obstacle, expressed in the world frame. \tabularnewline
				\hline 
				$g$&$g\approx9.81$~\accelunits{}.\tabularnewline
				\hline 
				$\boldsymbol{\xi}$ & Relative acceleration, expressed in the world frame: $\arraycolsep=2.1pt\boldsymbol{\xi}:=\left[\begin{array}{ccc}
					\mathbf{a}_{x} & \mathbf{a}_{y} & \mathbf{a}_{z}+g\end{array}\right]^{T}$. We will assume $\boldsymbol{\xi}\neq\boldsymbol{0}$.\tabularnewline
				\hline 
				
				\vspace{0.05cm}
				$\arraycolsep=1.4pt\left[\begin{array}{cccc}
					q_{w} & q_{x} & q_{y} & q_{z}\end{array}\right]^{T}$ & Components of a unit quaternion.\tabularnewline
				\hline 
				\vspace{0.6cm}
				$\psi$ & Angle such that %
				$\arraycolsep=1.2pt\boldsymbol{q}_{b}^{w}=\frac{1}{\sqrt{2(1+\bar{\boldsymbol{\xi}}_{z})}}\left[\begin{array}{c}
					1+\bar{\boldsymbol{\xi}}_{z}\\
					-\bar{\boldsymbol{\xi}}_{y}\\
					\bar{\boldsymbol{\xi}}_{x}\\
					0
				\end{array}\right]\circ\left[\begin{array}{c}
					c_{\psi/2}\\
					0\\
					0\\
					s_{\psi/2}
				\end{array}\right]$  (\cite{watterson2020control,tordesillas2021panther}).\tabularnewline
				\hline

				World frame~($w$), body frame~($b$)  and camera frame~($c$) &\vspace{0.2cm} \includegraphics[width=0.9\columnwidth]{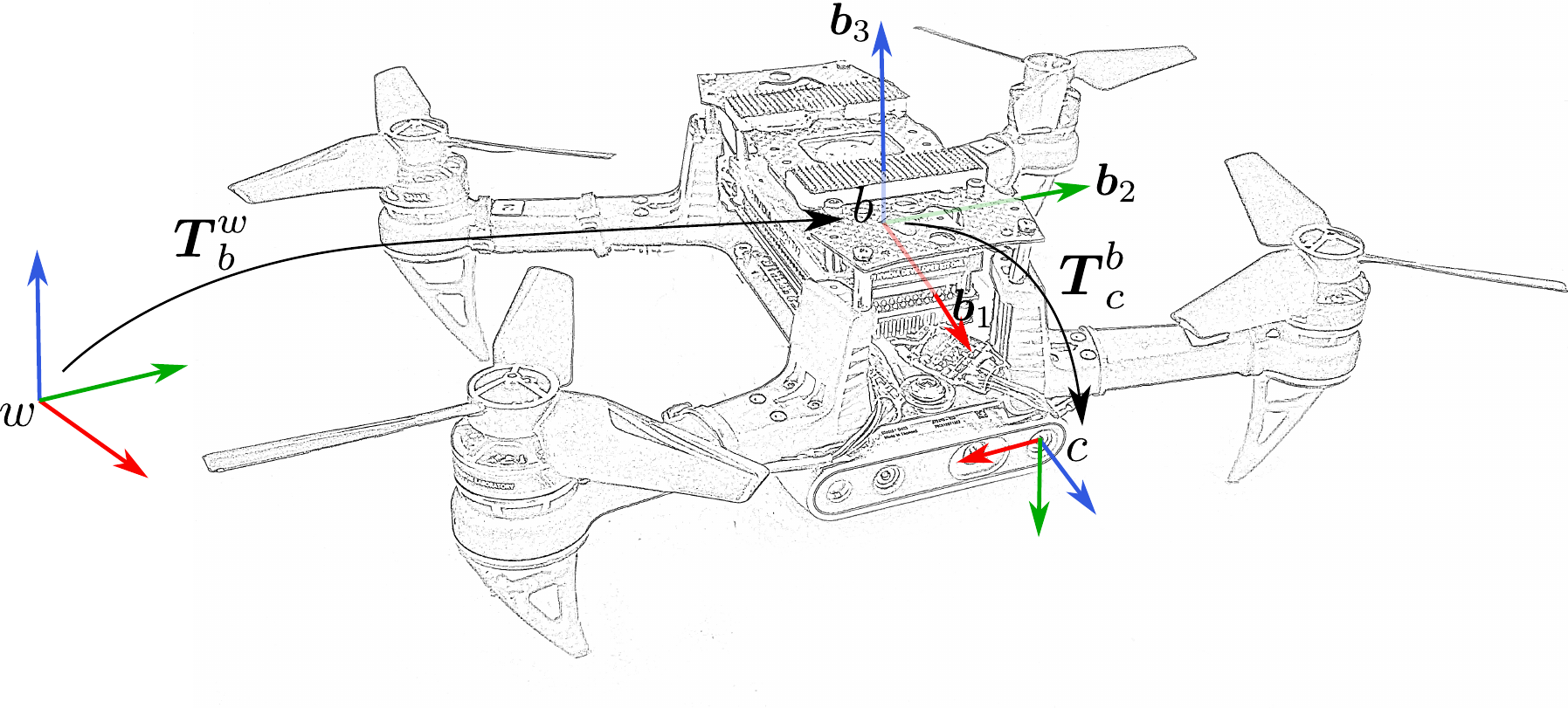}\\
				$\arraycolsep=1.4pt \boldsymbol{R}^w_b:=\left[\begin{array}{ccc}
					\boldsymbol{b}_1 & \boldsymbol{b}_2 & \boldsymbol{b}_3\end{array}\right]$. \\ $\boldsymbol{b}_3=\left(\boldsymbol{\xi}\right)_n$ due to the perpendicularity of the total thrust with respect to the plane spanned by $\boldsymbol{b}_1$ and $\boldsymbol{b}_2$.
				\tabularnewline
				\hline 
				Frame $f$  & Coordinate frame such that $\boldsymbol{t}^b_f=\boldsymbol{0}$, $\arraycolsep=2pt\boldsymbol{R}^w_f\boldsymbol{e}_z=\left[\begin{array}{ccc}
					0 & 0 & 1\end{array}\right]^{T}$ %
				, and that has the same $\psi$ as the frame $b$.\tabularnewline
				\hline 	
				
				$\mathbf{v}^f,\mathbf{a}^f$ & Velocity and Acceleration of the body w.r.t. the world frame, and expressed in the
				frame $f$.  $\in\mathbb{R}^{3}$.\tabularnewline
				\hline
				
				$\theta$ & Opening angle of the cone that approximates the FOV.\tabularnewline
				\hline 
				
				$n$ ($n_{\mathbf{p}}$ and $n_{\psi}$ ) & $n:=m-p-1$.
				
				$n+1$ is the number of control points of the spline.\tabularnewline
				\hline 
				
				$L_{\mathbf{p}}$, $L_{\psi}$ & $L_{\mathbf{p}}:=\{0,1,...,n_{\mathbf{p}}\}$, $L_{\psi}:=\{0,1,...,n_{\psi}\}$.\tabularnewline
				\hline 
				$l$ & Index of the control point. $l\in L_{\mathbf{p}}$
				for $\mathbf{p}(t)$, $l\in L_{\psi}$ for $\psi(t)$.\tabularnewline
				\hline 
				$\boldsymbol{q}_{l},\psi_{l}$ & Position B-Spline control point~$\left(\in\mathbb{R}^{3}\right)$,
				$\psi$ B-Spline control point~$\left(\in\mathbb{R}\right)$. \tabularnewline

				\hline
				$\poscpts{}$, $\poscptscropped{}$ &
				
				$\poscpts{}:=\{\left(\boldsymbol{q}_{l}\right)^f\}_{l\in L_{\mathbf{p}}}$. In other words, the position B-Spline control points of the planned trajectory for the UAV, expressed in frame $f$.
				
				$\poscptscropped{}:=\{\left(\boldsymbol{q}_{l}\right)^f\}_{l\in\{3,...,n_{\mathbf{p}}-2\}}$.

				\quad \includegraphics[width=0.97\columnwidth]{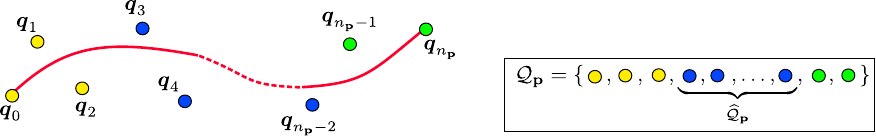}
				\tabularnewline
				
				\hline 	
				$\yawcpts{}$&$\{\psi_{l}\}_{l\in L_{\psi}}$. In other words, the $\psi$ B-Spline control points (with respect to frame $f$) of the planned trajectory for the UAV.   \tabularnewline
				\hline 
				\poscptsObst{} & B-Spline control points of a spline fit to the future predicted trajectory of obstacle. The future predicted trajectory of the obstacle can be obtained using \addrevision{a prediction module as in~\cite{tordesillas2021panther}}.\tabularnewline
				\hline 
				$\sizeObst{}$  & Length of each side of the axis-aligned bounding box of the obstacle. $\in \mathbb{R}^3$. \tabularnewline
				\hline 
				\addrevision{$\boldsymbol{s}_\text{UAV}$}  & \addrevision{Length of each side of the axis-aligned bounding box of the UAV. $\in \mathbb{R}^3$.} \tabularnewline
				\hline
				$T_{\text{pred}}$ & Prediction time for the future trajectory of the obstacle(s).\tabularnewline
				\hline
				$n_s$ & Number of trajectories produced by the student. It is a user-chosen parameter, and it is fixed (i.e., does not change between replanning steps). \tabularnewline
				\hline
				$n_\text{runs}$, $n_\text{sols}$, $n_e$ & The optimization problem of the expert is run $n_\text{runs}$ times (with different initial guesses), producing $n_\text{sols}\le n_\text{runs}$ distinct trajectories. The trajectories produced by the expert are then the best $n_e=\text{min}(n_\text{sols},n_s)$ trajectories obtained. \tabularnewline
				\hline
				\multicolumn{2}{|l|}{Snapshot at $t=t_{1}$ (current time):}\tabularnewline
				\multicolumn{2}{|c|}{\includegraphics[width=0.6\columnwidth]{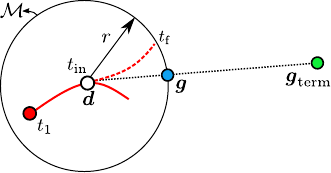}}\tabularnewline
				\multicolumn{2}{|l|}{$\boldsymbol{g}_{\text{term}}$ (\tikzcircle[black,fill=green]{15pt})
					is the terminal goal, and \tikzcircle[black,fill=red]{15pt} is the
					current position of the UAV.}\tabularnewline
				\multicolumn{2}{|l|}{\CurrTrajDeepPanther \ is the trajectory the UAV is currently executing.}\tabularnewline
				\multicolumn{2}{|l|}{\NextTrajDeepPanther \ is the trajectory the UAV is currently planning, $t\in\left[t_{\text{in}}, t_{\text{f}}\right]$. } \tabularnewline
			\multicolumn{2}{|l|}{$\boldsymbol{d}$ (\tikzcircle[black,fill=white]{15pt}) is a point
				in \CurrTrajDeepPanther, used as the initial position of \NextTrajDeepPanther.}\tabularnewline
			\multicolumn{2}{|l|}{$\mathcal{M}$ is a sphere of radius $r$ around $\boldsymbol{d}$.}\tabularnewline
			\multicolumn{2}{|l|}{$\boldsymbol{g}$ (\tikzcircle[black,fill=blue_light]{15pt}) is the
				projection of $\boldsymbol{g}_{\text{term}}$ (\tikzcircle[black,fill=green]{15pt}) onto the sphere $\mathcal{M}$.}\tabularnewline
			\multicolumn{2}{|l|}{$T$ is the total time of the planned trajectory. I.e., $T:=t_{\text{f}}-t_{\text{in}}$.}\tabularnewline
			\hline  				
		\end{tabular}
		\par\end{centering}
}
\end{table}

\section{Deep-PANTHER}\label{sec:deep_panther_section}

Deep-PANTHER is a multimodal trajectory planner able to generate a trajectory that avoids a dynamic obstacle, while trying to keep it in the FOV. To achieve very fast computation times, we leverage imitation learning, where Deep-PANTHER is the student (a neural network) that is trained to imitate the position trajectories generated by an optimization-based expert (Section~\ref{sec:expert_and_student}). Both the student and the expert have an observation as input and an action as output (Section~\ref{sec:observation_and_action}). The multimodality is captured through the design of the loss function (Section~\ref{sec:loss_multiple_hypothesis}), and the trajectories for the extra degree of freedom of the rotation ($\psi$) can then be obtained from the position trajectories (Section~\ref{sec:closed_form_yaw}). The final trajectory chosen for execution is obtained according to the cost and the constraint satisfaction (Section~\ref{sec:testing}). \addlastrevision{This paper uses the notation shown in Table~\ref{tab:Notation_deep_panther}.}

\subsection{Expert and Student}\label{sec:expert_and_student}

Our prior work~\cite{tordesillas2021panther} developed PANTHER, an optimization-based perception-aware trajectory planner able to avoid dynamic obstacles while keeping them in the FOV. However, and as discussed in Section~\ref{sec:introduction_deep_panther}, real-time computation was achieved at the expense of conservative solutions. Hence, we design \pantherStar{} (the expert) by reducing the conservativeness of PANTHER as follows:
\begin{itemize}
	\item \addrevision{The planes that separate the trajectory of the UAV from the obstacles \cite{tordesillas2021panther}} and the total time of the planned trajectory $T$ are included as decision variables. To ensure that $T$ does not go beyond the prediction horizon, the constraint $0\le T\le T_{\text{pred}}$ is imposed for both the expert and the student. Here, $T_{\text{pred}}$ is the total time of the future predicted trajectory of the obstacle, and it is a user-chosen parameter.
	\item The future predicted trajectory of the obstacle is a spline whose control points are \poscptsObst{}.
	\item The optimization problem is run $n_\text{runs}$ times (with different initial guesses obtained by running the OSA~\cite{tordesillas2020mader}), and $n_\text{sols}\le n_\text{runs}$ distinct trajectories are obtained. %
\end{itemize}
The student (Deep-PANTHER) consists of a fully connected feedforward neural network with two hidden layers, 64 neurons per layer, and with the ReLU activation function. The student produces a total of $n_s$ trajectories, where $n_s$ is a user-chosen parameter. Note that the trajectories produced by the expert are then the best $n_e=\text{min}(n_\text{sols},n_s)$ trajectories obtained in the optimization.

\newcommand{\mye}{ \ensuremath{\posObstacleWorldFrame{} - \posBodyWorldFrame{}}}

\begin{figure*}
	\begin{centering}
		\includegraphics[width=1\textwidth]{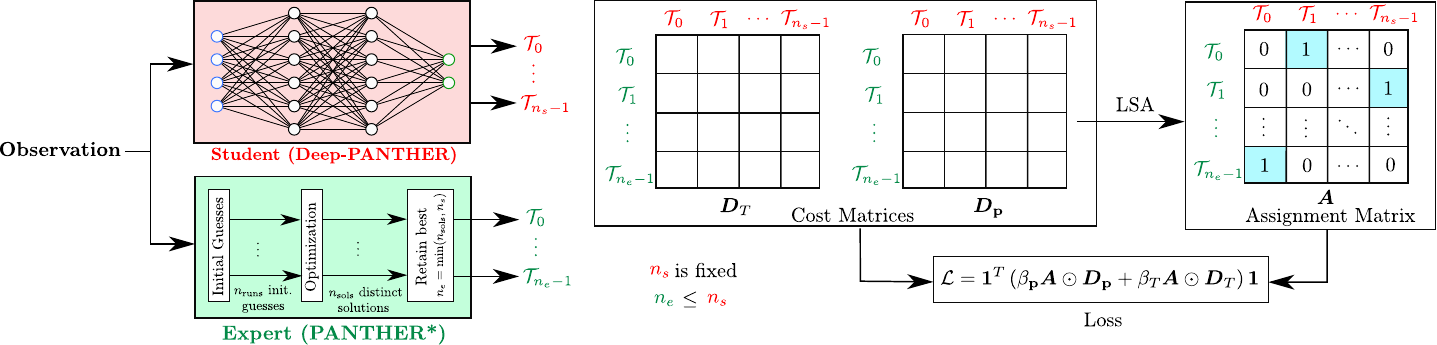}
		\par\end{centering}
	\caption[Multimodal training in Deep-PANTHER]{Multimodal training in Deep-PANTHER. The student outputs a fixed number of trajectories, denoted as~$n_s$. The expert (\addrevision{\pantherStar{}}) produces $n_e$ trajectories, where $n_e\le n_s$. Then, the cost matrix in position space ($\boldsymbol{D}_{\mathbf{p}}$) and in time space ($\boldsymbol{D}_{T}$) are computed. Using $\boldsymbol{D}_{\mathbf{p}}$, the linear sum assignment (LSA) problem is solved 
	to find the assignment matrix~$\boldsymbol{A}$, which is then used in the loss to penalize the expert-student assigned pairs. 
		\label{fig:hung} } 
		\vskip-2.3ex
\end{figure*}

\newcommand{\RedPsiTraj}[2][]{\tikz[baseline=0.0ex]\draw [red,thick] (0,0.08) -- (0.5,0.08);}
\begin{figure*}
	\begin{centering}
		\includegraphics[width=1\textwidth]{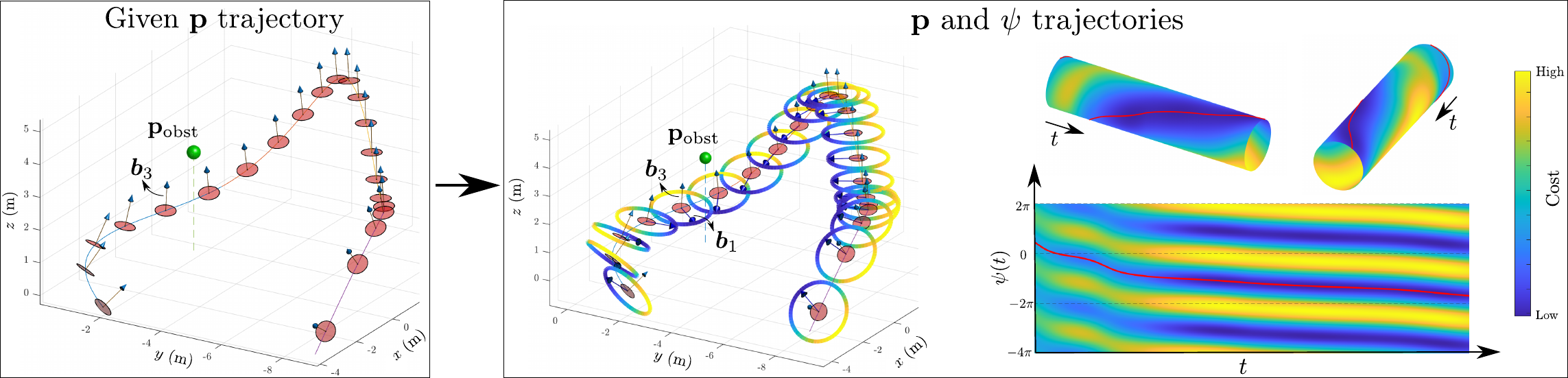}
		\par\end{centering}
	\caption[Optimal $\psi$ trajectory given the position trajectory]{Optimal $\psi$ trajectory (\RedPsiTraj{} in the right plots) given the position trajectory defined by $\left( \left(\poscpts{}\right)_k, T_k  \right)$. A spline is then fit to this $\psi$ trajectory found. The sphere \tikzcircle[black,fill=green]{25pt} denotes the position of the obstacle~\posObstacleWorldFrame{}. For visualization purposes, we show here a static obstacle, but this method is also applicable when the obstacle is dynamic. \label{fig:closed_form_yaw}}
		\vskip-2.3ex
\end{figure*}

\subsection{Observation and Action}\label{sec:observation_and_action}

We use the observation
$\left(   \mathbf{v}^{f},   \mathbf{a}^{f},  \boldsymbol{g}^{f},  \addrevision{\dot{\psi}}    , \poscptsObst{}, \sizeObst{} \right)$, 
where, as defined in Table~\ref{tab:Notation_deep_panther},  $\mathbf{v}^f\addrevision{\in\mathbb{R}^3}$, $\mathbf{a}^f\addrevision{\in\mathbb{R}^3}$, $\boldsymbol{g}^f\addrevision{\in\mathbb{R}^3}$, and \poscptsObst{} are, respectively, the velocity of the UAV, the acceleration of the UAV, the projection of the terminal goal $\boldsymbol{g}_{\text{term}}$, and the control points of a spline fit to the future predicted trajectory of the obstacle. All of these quantities are expressed in the frame~$f$. \addrevision{$\dot{\psi}\in\mathbb{R}$ is the derivative of $\psi(t)$}. $\sizeObst{}\in\mathbb{R}^3$ contains the length of each side of the axis-aligned bounding box of the obstacle. In this work, we use a spline in $\mathcal{S}^3_{3,13}$ for the predicted trajectory of the obstacle, \addrevision{which means that \poscptsObst{} contains 10 position control points, each one in~$\mathbb{R}^3$. This leads to an} observation size of $43$.

The action is given by $(\mathcal{T}_k)_{k\in\{0,\hdots,\beta-1\}}$, where $\beta=n_s$ for the student, and $\beta=n_e$ for the expert, and where $\mathcal{T}_k:=\left( \left(\poscptscropped{}\right)_k, T_k  \right)\;.$
As defined in Table~\ref{tab:Notation_deep_panther}, $\poscptscropped{}$ contains all the B-Spline control points of the planned trajectory expressed in frame~$f$ except the first three and the last two, while $T$ is the total time of the planned trajectory. Note that the first three and the last two control points need not to be included in the action because they are determined directly from the total time $T$ and the initial and final conditions. We model the planned trajectories (for both the expert and the student) as splines in $\mathcal{S}^3_{3,12}$, leading to an action size of $13\beta$. The relationship between $n_s$ and $n_e$ is explained in Section~\ref{sec:expert_and_student} and Table~\ref{tab:Notation_deep_panther}.

The key advantage of using \poscptscropped{} instead of \poscpts{} is that every trajectory generated by the student will satisfy by construction the initial and final conditions for any given observation. It also helps reduce the action size. Moreover, the advantage of using the B-Spline position control points, instead of sampled future positions as in~\cite{loquercio2021learning}, is that every trajectory generated by the student is smooth by construction ($\mathcal{C}^2$-continuous in our case), %
and it also avoids the need of a post-projection step into polynomial space.

\newcommand{\summaryRWTAMSEcomparison}{Compared to RWTAr, our approach achieves an average MSE between 1.09 and 18.02 times smaller. Compared to RWTAc, our approach achieves an average MSE between 2.35 and 2.68 times smaller}

\newcommand{\explanationcaserzero}{For instance, the case $\kappa=0$ corresponds to the trajectory of the student that best predicts an expert trajectory}

\subsection{Loss: Capturing Multimodality}\label{sec:loss_multiple_hypothesis}

As discussed in Section~\ref{sec:introduction_deep_panther} and Fig.~\ref{fig:modes_expert}, the number of trajectories found by the expert changes depending on the specific observation. 
To train a neural network with a fixed-size output ($n_s$ trajectories) to predict the varying-size output of the expert ($n_e$ trajectories),
we propose to use the approach shown in Fig.~\ref{fig:hung}. The observation is passed through the neural network of the student to generate $n_s$ trajectories, and through the expert to produce $n_e$ trajectories. We then define $\boldsymbol{D}_{\mathbf{p}}$ as a matrix whose element $(i,j)$ is the mean squared error (MSE) between the position control points of the $i$-th trajectory of the expert and the position control points of the $j$-th trajectory of the student. A similar definition applies to $\boldsymbol{D}_{T}$, but using the total time of the trajectory instead of the control points.

Letting $\boldsymbol{A}$ denote the assignment matrix (whose $(i,j)$ element is 1 if the $i$-th trajectory of the expert has been assigned to the $j$-th trajectory of the student, and 0 otherwise), we then find the optimal $\boldsymbol{A}$ that minimizes the assignment cost $\boldsymbol{1}^T\left(\boldsymbol{A}\odot\boldsymbol{D}_{\mathbf{p}}\right)\boldsymbol{1}$, and that assigns a distinct student trajectory to every expert trajectory. 
Here, $\odot$ denotes the element-wise product and $\boldsymbol{1}$ is a column vector of ones. This is an instance of the linear sum assignment (LSA) problem, and the optimal  $\boldsymbol{A}$ can be obtained leveraging the Jonker-Volgenant algorithm~\cite{crouse2016implementing} (a variant of the Hungarian algorithm~\cite{kuhn1955hungarian}).
As we have that $n_e\le n_s$, all the rows of $\boldsymbol{A}$ sum up to~1, $n_e$ columns sum up to~1, and $(n_s-n_e)$ columns sum up to~0. To penalize only the MSE of the optimally-assigned student-expert pairs, the loss is then computed as
$$\mathcal{L}=\boldsymbol{1}^{T}\left(\beta_{\mathbf{p}}\boldsymbol{A}\odot\boldsymbol{D}_{\mathbf{p}}+\beta_{T}\boldsymbol{A}\odot\boldsymbol{D}_{T}\right)\boldsymbol{1}\;,$$
where $\beta_{\mathbf{p}}$ and $\beta_{T}$ are user-chosen weights.

Our approach ensures that all the expert trajectories have exactly one distinct student trajectory assigned to them, see Fig~\ref{fig:wta_rwta_ours_explanation}. %
Compared to WTAr, RWTAr, WTAc, and RWTAc, our LSA loss prevents the same student trajectory from being assigned to several expert trajectories (reducing therefore the equilibrium issues), guarantees that all the trajectories of the expert are captured in every training step, and also prevents the same expert trajectory from having several student trajectories assigned to it (being therefore less prone the mode collapse problems).

\newcommand{\TkToPosTraj}{Then, for each $\mathcal{T}_k$, the initial and final conditions are imposed to generate the position trajectory, defined by all the position control points $(\poscpts{})_k$ and the total time $T_k$}

\begin{figure*}
	\begin{centering}
		\includegraphics[width=1\textwidth]{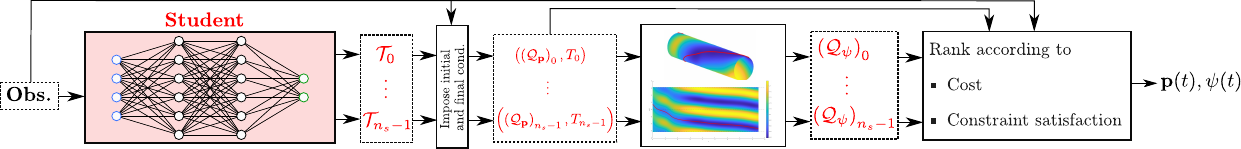}
		\par\end{centering}
	\caption[Generation of $\mathbf{p}(t)$ and $\psi(t)$ from the observation]{Generation of $\mathbf{p}(t)$ and $\psi(t)$ from the observation. 
	 \label{fig:testing_time}
	}
	\vskip-3.0ex
\end{figure*}

\newcommand{\ourPolicy}{\ensuremath{\pi_{\text{LSA}}}}

\subsection{Generation of $\psi$ given the Position Trajectory}\label{sec:closed_form_yaw}

Each $\mathcal{T}_k$, together with the initial and final conditions contained in the observation, defines the position trajectory. 
Since $\boldsymbol{b}_3:=\boldsymbol{R}^w_b\boldsymbol{e}_z=\left(\boldsymbol{\xi}\right)_n$ (see Fig.~\ref{fig:closed_form_yaw} and Table~\ref{tab:Notation_deep_panther}), this position trajectory determines part of the rotation, but leaves $\psi$ free. We now derive\footnote{For simplicity, here we focus on the case where $\tiny \arraycolsep=1.0pt\boldsymbol{R}_{c}^{b}=\left[\begin{array}{ccc}
		0 & 0 & 1\\
		-1 & 0 & 0\\
		0 & -1 & 0
	\end{array}\right]$ and $\boldsymbol{t}_{c}^{b}=\boldsymbol{0}$. A similar derivation applies to more general cases. \addrevision{See also \cite{falanga2017aggressive}.}} a closed-form expression for $\psi(t)$ that maximizes the presence of the obstacle in the FOV given the position trajectory. Let $\posBodyWorldFrame{}:=\boldsymbol{t}^w_b$ be the position of the UAV, and let \posObstacleWorldFrame{} denote the position of the obstacle (both expressed in the world frame). Let us also define $\boldsymbol{b}_{1}:=\boldsymbol{R}^w_b \boldsymbol{e}_x$.  Using a cone with opening angle $\theta$ to model the FOV, the obstacle is in the FOV if and only if $\text{cos}(\theta/2)\le\boldsymbol{b}_{1}^{T}\left( \mye{} \right)_n$. We can therefore maximize the presence of the obstacle in the FOV by solving the following optimization problem:
$$
\underset{\boldsymbol{b}_{1}}{\min}-\boldsymbol{b}_{1}^{T}\left( \mye{} \right)_n \text{ subject to }   \boldsymbol{b}_{1}^{T}\boldsymbol{\xi}=0 \text{ and }  \left\Vert \boldsymbol{b}_{1}\right\Vert ^{2}=1 
$$
where the two constraints guarantee that $\boldsymbol{b}_1$ is a unit vector perpendicular to $\boldsymbol{\xi}$. Computing the Lagrangian and solving the Karush-Kuhn-Tucker (KKT) conditions~\cite{kuhn1951nonlinear, karush1939minima} yields the optimal solution:\footnote{Note that Eq.~\ref{eq:closed_form_yaw} presents a singularity when $\left(\mye{}\right)$ is parallel to~$\boldsymbol{\xi}$. In that case, we can choose any $\boldsymbol{b}_1$, since all of them are perpendicular to $\left(\mye{}\right)$ and therefore achieve the same (zero) cost in the objective function.}
\begin{equation}\label{eq:closed_form_yaw}
	\boldsymbol{b}_{1}=\left(\left( \mye{} \right)-\frac{\left( \mye{} \right)^{T}\boldsymbol{\xi}}{\left\Vert \boldsymbol{\xi}\right\Vert ^{2}}\boldsymbol{\xi}\right)_{n}
\end{equation}

Given that \posObstacleWorldFrame{}, \posBodyWorldFrame{}, and~$\boldsymbol{\xi}$ 
are functions of time, Eq.~\ref{eq:closed_form_yaw} gives the evolution of $\boldsymbol{b}_1$ that maximizes the presence of the obstacle in the FOV (see Fig.~\ref{fig:closed_form_yaw}). 
$\psi(t)$ can then be easily obtained from $\boldsymbol{b}_1$ and $\boldsymbol{b}_3$, and a spline is fit to it to obtain the control points~\yawcpts{}.

Note that, in \pantherStar{}, position and rotation are coupled together in the optimization~\cite{tordesillas2021panther}. This coupling helps reduce the conservativeness that arises when they are optimized separately~\citePADec{}. Deep-PANTHER (the student) learns to predict the position trajectory resulting from this \textit{coupled} optimization problem, and then the closed-form solution is leveraged to obtain $\psi$ from this position trajectory. In other words, Deep-PANTHER benefits from the coupling (since it is learning one of the outputs of the coupled optimization problem), while leveraging the closed-form solution for $\psi$. 

\subsection{Testing}\label{sec:testing}

In testing time the procedure is as follows (see Fig.~\ref{fig:testing_time}): The observation is fed into the neural network, which produces $(\mathcal{T}_k)_{k\in\{0,\hdots,n_s-1\}}$ (i.e., the intermediate position control points and the total times). \TkToPosTraj{}. The optimal $\psi$ control points $\left(\yawcpts{}\right)_k$ are then obtained as explained in Section~\ref{sec:closed_form_yaw}. Then, and using the observation, each triple $\left((\poscpts{})_k, \left(\yawcpts{}\right)_k, T_k\right)$ is ranked according to the cost and the constraint satisfaction. The trajectory chosen for execution is then the one that is collision-free and achieves the smallest augmented cost, which is defined as $c_\text{obj}+\lambda c_\text{dyn lim}$, where $c_\text{obj}$ is the cost of the \addrevision{objective function,\footnote{\addrevision{In this paper, PANTHER, \pantherStar{}, and Deep-PANTHER use the following cost:
$c_\text{obj}=\alpha_{\mathbf{{j}}}\int_{0}^{T}\left\Vert \mathbf{j}\right\Vert ^{2}dt+\alpha_{\psi}\int_{0}^{T}\left(\ddot{\psi}\right)^{2}dt-\alpha_{\text{FOV}}\int_{0}^{T}\left(\text{inFOV}(\boldsymbol{T}_{c}^{w},\mathbf{p}_{\text{obs}})\right)^3 dt+\alpha_{\boldsymbol{g}}\left\Vert \mathbf{p}(t_{f})-\boldsymbol{g}\right\Vert^{2} + \alpha_T T$ , where    $\left\{\alpha_{\mathbf{{j}}},\;\alpha_{\psi},\;\alpha_{\text{FOV}},\;\alpha_{\boldsymbol{g}},\;\alpha_T\right\}$ are nonnegative weights.}}} $c_\text{dyn lim}$ is a soft cost that penalizes the velocity, acceleration, and jerk violations, and $\lambda>0$.  If none of the trajectories generated by the student are collision-free, the UAV will continue executing the trajectory it had in the previous replanning step (which is collision-free) and will replan again.

\begin{figure}
	\begin{centering}
		\includegraphics[width=1\columnwidth]{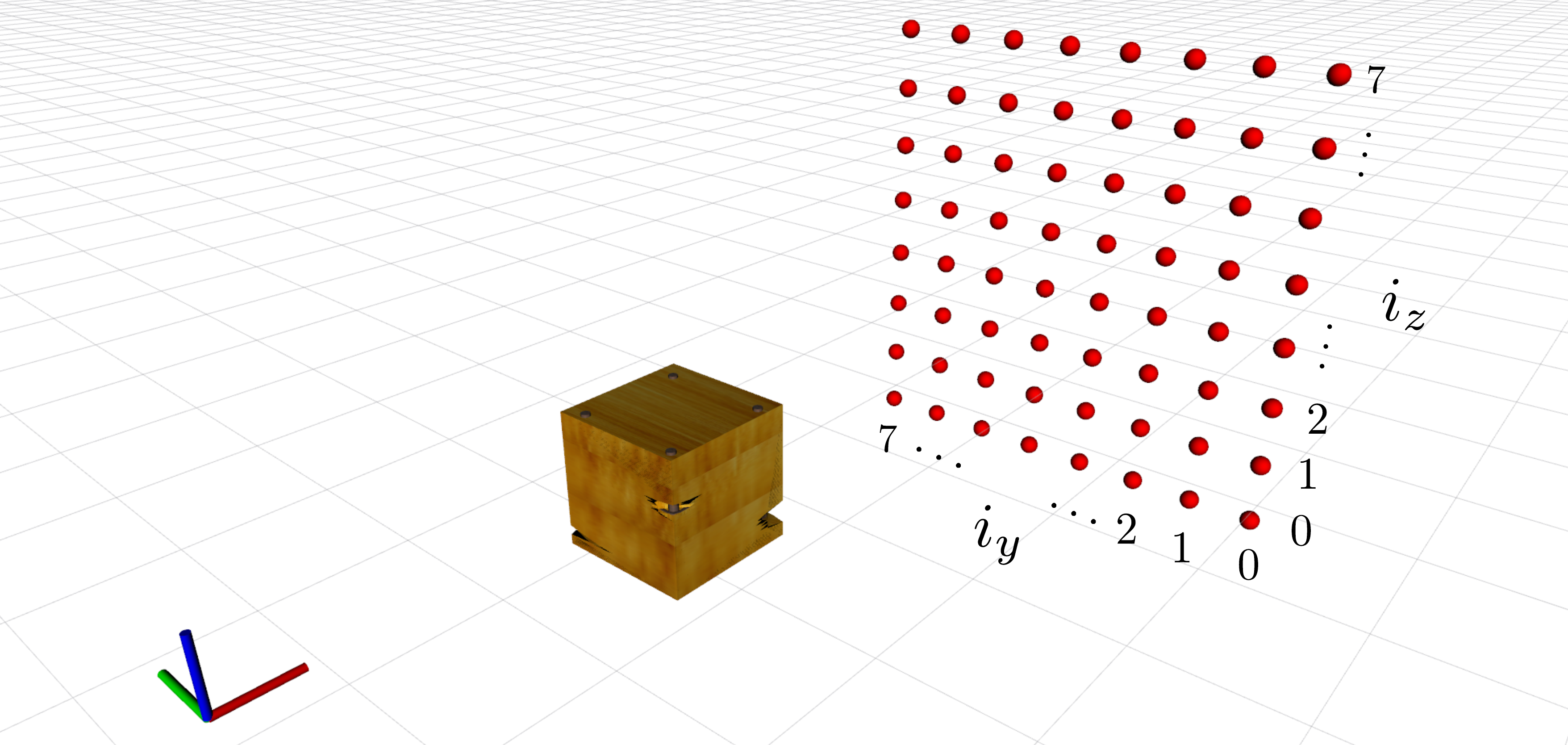}
		\par\end{centering}
	\caption[Testing scenario with a static obstacle and 64 different goals]{
		Testing scenario consisting of a static obstacle located at $\arraycolsep=2.0pt
		\posObstacleWorldFrame{} = \left[\begin{array}{ccc}
			2.5 & 0 & 1\end{array}\right]^T
		$~m and $64$ different $\boldsymbol{g}_{\text{term}}=\arraycolsep=3.0pt\left[\begin{array}{ccc}
			7 & a & 1+b\end{array}\right]^T$~m (\tikzcircle[black,fill=red]{25pt} in the figure), where $a$ and $b$ are evenly spaced in $[-1.7,1.7]$~m. The initial location (coordinate frame in the figure) is $\arraycolsep=2.0pt\left[\begin{array}{ccc}
			0 & 0 & 1\end{array}\right]^T$~m.
 \label{fig:comparison_safe_trajs_rviz}
		}
		\vskip 2.5ex

	\begin{centering}
		\includegraphics[width=1\columnwidth]{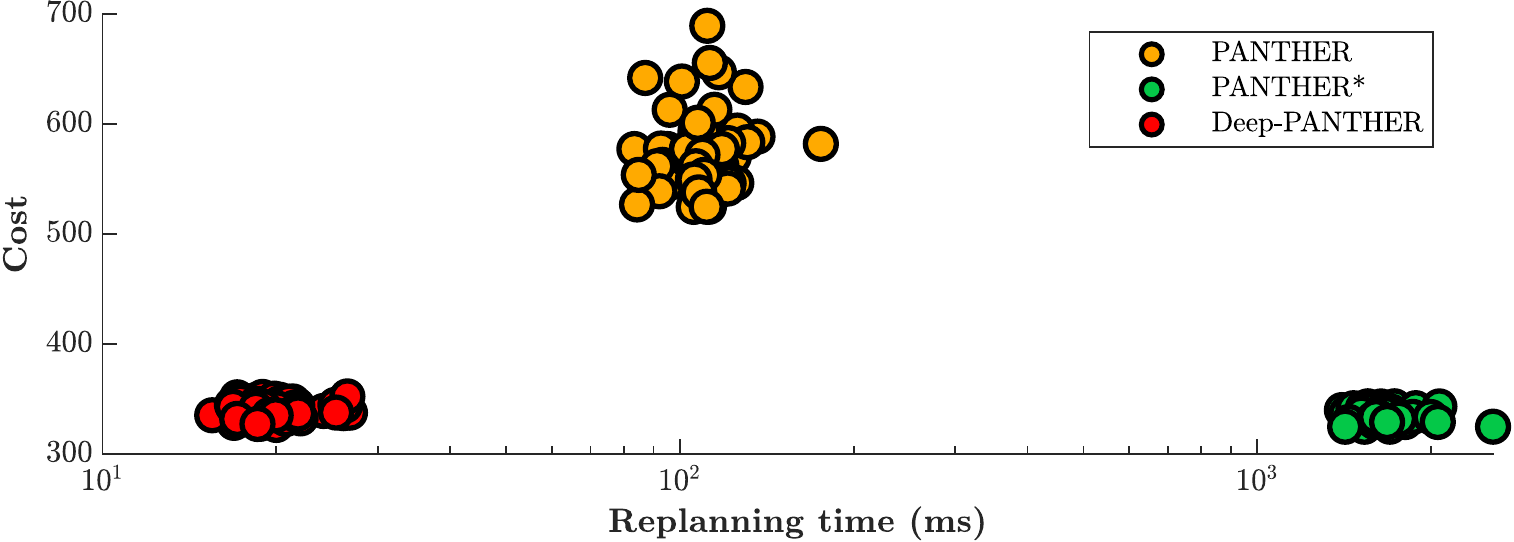}
		\par\end{centering}
	\caption[Comparison of the cost and replanning time]{Comparison of the cost and replanning time. 
		Deep-PANTHER is able to obtain a similar cost to the one obtained by \pantherStar{}, but with a computation time two order of magnitude smaller. Compared to PANTHER, Deep-PANTHER is able to get a smaller cost, and one order of magnitude faster. \addrevision{The cost is defined in Section~\ref{sec:testing}}. Note the logarithmic scale on the $x$ axis.  
		\label{fig:cost_vs_comp_time_log_scale}}
		\vskip-2.5ex
\end{figure}

\begin{figure*}
	\definecolor{mygreenboxplot}{RGB}{27, 179, 31}
	\definecolor{myyellowboxplot}{RGB}{255, 187, 111}
	\newcommand{\YellowDashedTraj}[2][]{\tikz[baseline=0.0ex]\draw [myyellowboxplot,thick, dash pattern=on 3pt off 1pt] (0,0.08) -- (0.5,0.08);}
	\begin{centering}
		\includegraphics[width=1\textwidth]{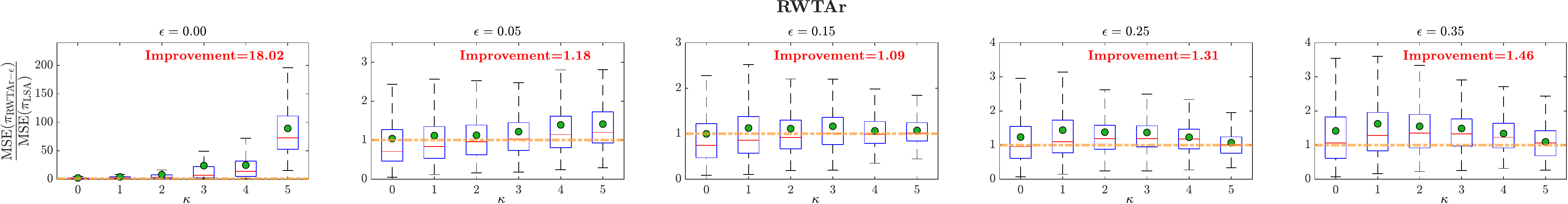}
		\vskip 0.1cm
		\includegraphics[width=1\textwidth]{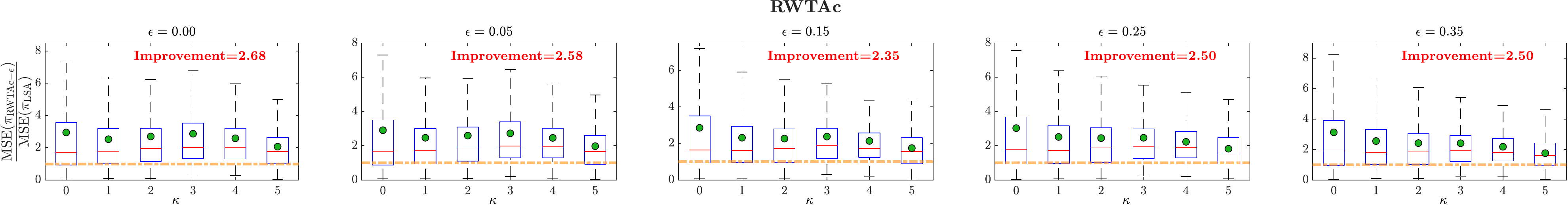}
		\par\end{centering}
	\caption[Comparison between the MSE loss of \ourPolicy{}, ${\pi}_{\text{RWTAr-$\epsilon$}}$, and ${\pi}_{\text{RWTAc-$\epsilon$}}$]{Comparison between the MSE loss of \ourPolicy{}, ${\pi}_{\text{RWTAr-$\epsilon$}}$, and ${\pi}_{\text{RWTAc-$\epsilon$}}$. In each of the boxplots, \tikzcircle[black,fill=mygreenboxplot]{25pt}~represents the mean. The dashed yellow line \YellowDashedTraj{} represents an MSE ratio of 1. %
	In the plots, $\kappa$ is the index of the ranking order based on the MSE loss with respect to the expert. 
	If $n_e<n_s=6$, that demonstration is not taken into account for the boxplots with $\kappa\ge n_e$.
		\label{fig:comparison_estimation_hypothesis_rwta}}
		\vskip-2.5ex
\end{figure*}

\section{Results and Discussion}\label{sec:results_and_discussion_deep_panther}

To better compare the different aspects of the proposed framework, Section~\ref{sec:static_planning_deep_panther} first focuses on a stopped UAV that needs to plan a trajectory from the start location to the goal (without moving along that planned trajectory) while avoiding a static obstacle. Then, Section~\ref{sec:replanning_dynamic_deep_panther} studies the more general case where a UAV is flying and constantly replanning in a dynamic environment.

We use $n_s=6$, $n_\text{runs}=10$, $T_{\text{pred}}=6$~s, and $\beta_{\mathbf{p}}=\beta_{T}=1$.\footnote{Note that $\beta_{\mathbf{p}}$ and $\beta_{T}$ are adimensional because $\boldsymbol{D}_{\mathbf{p}}$ and $\boldsymbol{D}_{T}$ are computed from normalized actions in~$[-1,1]$.}  %
To train the neural network we use the Adam optimizer~\cite{kingma2015adam} and a learning rate of $10^{-3}$. In all these simulations, and for all the algorithms tested, \poscptsObst{} is obtained by simply fitting a spline to the ground-truth future positions of the obstacle. In real-world applications, this future predicted trajectory of the obstacle \addrevision{can be obtained from past observations~\cite{tordesillas2021panther}}.

\addrevision{Note that the selection of $n_s$ and $n_\text{runs}$ sufficiently high helps reduce a potential bias problem that could appear if the expert generated very few (2 or 3) trajectories. Moreover, we also randomize the training environments to help reduce this potential bias (see following subsections).}

\subsection{Static Obstacle}\label{sec:static_planning_deep_panther}

In this section, the task is to plan once from the starting location to the goal (i.e., the UAV does not move along the planned trajectory and/or replan again). We collect $2$K (observation, expert action) pairs,\footnote{
	\addlastrevision{The code contains the details of the randomization performed.}
} and use 75\% of these pairs to train the student offline (the rest of the pairs are used as the evaluation dataset in the MSE comparisons of Section~\ref{sec:multi_hypothesis_results_panther}). Section~\ref{sec:comp_time_cost_deep_panther} first compares the cost vs replanning time, and then Section~\ref{sec:multi_hypothesis_results_panther} analyzes how well the multimodality is captured.

\begin{figure*}
	
	\definecolor{mylightgreen_deep_panther}{RGB}{203, 254, 203}
	\definecolor{mylightred_deep_panther}{RGB}{254, 203, 203}
	
	\begin{centering}
		\includegraphics[width=1\textwidth]{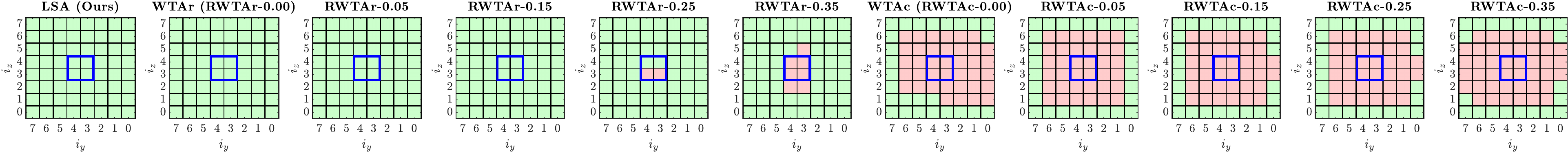}
		\par\end{centering}
	\caption[Comparison of the collision-free trajectories produced by LSA (our approach), RWTAr-$\epsilon$, and RWTAc-$\epsilon$]{Comparison of the collision-free trajectories produced by LSA (our approach), RWTAr-$\epsilon$, and RWTAc-$\epsilon$. Each cell in each square represents a different $\boldsymbol{g}_\text{term}$ ($i_y$ and $i_z$ are defined in Fig.~\ref{fig:comparison_safe_trajs_rviz}), where \tikzrectangleDeepPanther{black,fill=mylightgreen_deep_panther} means that at least one collision-free trajectory is obtained, while \tikzrectangleDeepPanther{black,fill=mylightred_deep_panther} means that none of the trajectories are collision-free. The blue square \tikzrectangleDeepPanther{blue,fill=white} is the projection of the obstacle onto the $y$-$z$ plane. 
	\label{fig:comparison_safe_trajs}}
	\vskip-3.5ex
\end{figure*}
\begin{figure}
	\definecolor{lightblue}{RGB}{0,215,255}
	\begin{centering}
		\includegraphics[width=1\columnwidth]{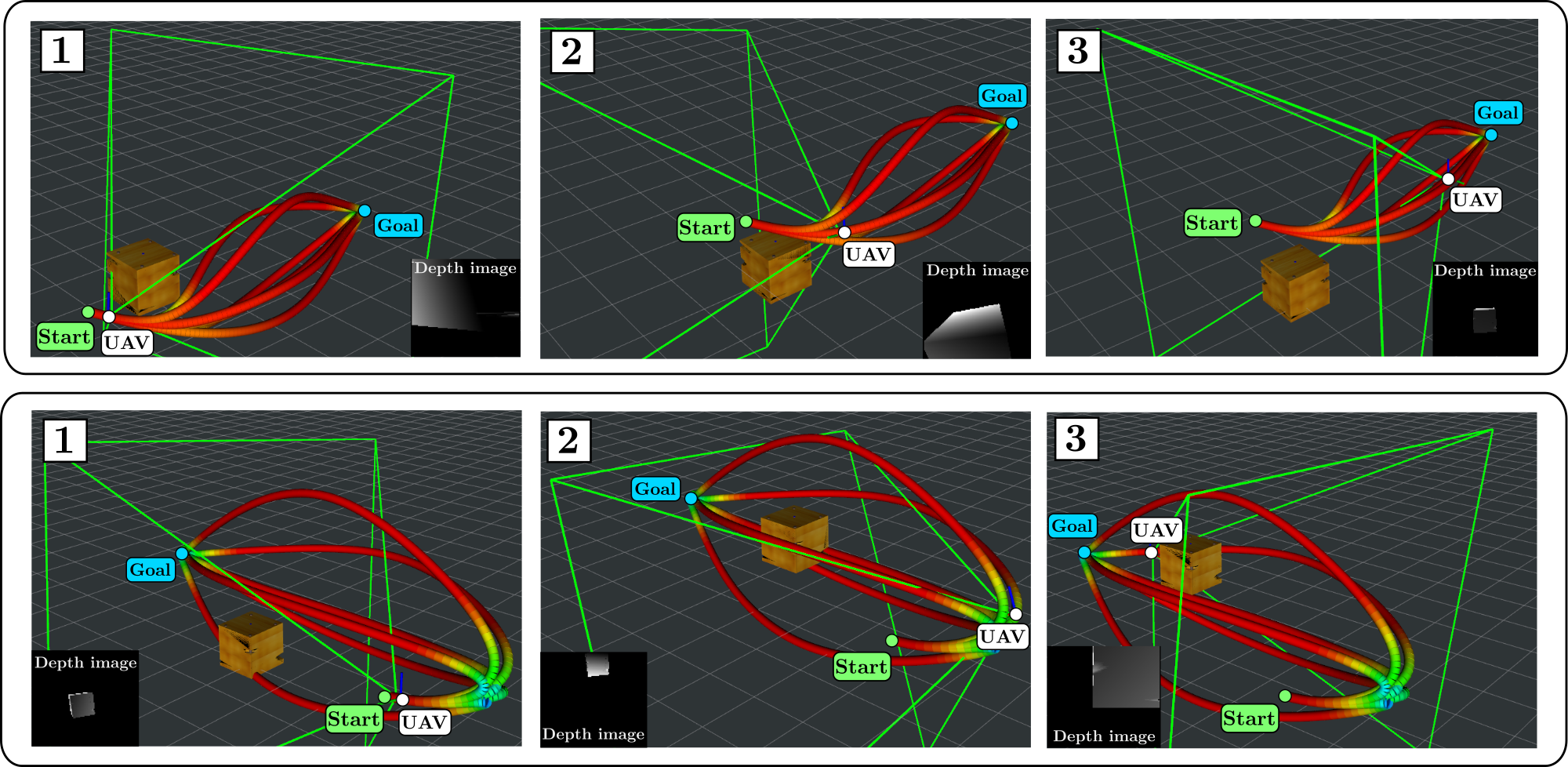}
		\par\end{centering}
	\caption[Snapshots of the trajectories produced by Deep-PANTHER in a dynamic environment]{Snapshots of the trajectories produced by Deep-PANTHER in a dynamic environment. \addlastrevision{The colormap represents the velocity (red denotes a higher velocity).}
		\label{fig:snapshots_dyn_obs}}
		\vskip-3.3ex
\end{figure}

\subsubsection{Cost vs Replanning \addrevision{T}ime}\label{sec:comp_time_cost_deep_panther}
We compare the cost vs replanning time of these three different approaches: PANTHER (Ref.~\cite{tordesillas2021panther}), \pantherStar{} (the expert, see Section~\ref{sec:expert_and_student}) and Deep-PANTHER (the student). The testing environment is shown in Fig.~\ref{fig:comparison_safe_trajs_rviz}. For \pantherStar{} and Deep-PANTHER (which generate a multimodal output), we use in the comparisons the best (i.e., with smallest cost) collision-free trajectory found.
The results are shown in Fig.~\ref{fig:cost_vs_comp_time_log_scale}, which highlights that Deep-PANTHER obtains a total cost similar to the one obtained by \pantherStar{}, but with a computation time that is two orders of magnitude smaller. Compared to PANTHER, Deep-PANTHER is able to obtain a lower cost, and with an improvement of one order of magnitude in computation time. For each of the 64 simulations performed, the trajectory obtained by all the algorithms is collision-free.

\subsubsection{Multimodality}\label{sec:multi_hypothesis_results_panther}

Let \ourPolicy{} denote the policy trained using the approach presented in Section~\ref{sec:loss_multiple_hypothesis}.
As explained in Section~\ref{sec:introduction_deep_panther}, another possible approach is to use ${\text{RWTAr}}$~\cite{makansi2019overcoming, rupprecht2017learning}, where the assignment matrix $\boldsymbol{A}$ has the value $(1-\epsilon)$ in the minimum elements of each row of $\boldsymbol{D}_{\mathbf{p}}$, and $\frac{\epsilon}{n_{s}-1}$ elsewhere. Similarly, ${\text{RWTAc}}$~\cite{loquercio2021learning} uses an assignment matrix $\boldsymbol{A}$ that has the value $(1-\epsilon)$ in the minimum elements of each column of $\boldsymbol{D}_{\mathbf{p}}$, and $\frac{\epsilon}{n_{e}-1}$ elsewhere.
 A policy trained using these approaches for a given $\epsilon\ge0$ will be denoted as ${\pi}_{\text{RWTAr-$\epsilon$}}$ and ${\pi}_{\text{RWTAc-$\epsilon$}}$. Note that ${\text{WTAr}\equiv\text{RWTAr}}$ and ${\text{WTAc}\equiv\text{RWTAc}}$ when $\epsilon=0$.
We first train 11 policies  (\ourPolicy{}, ${\pi}_{\text{RWTAr-$\epsilon$}}$, and ${\pi}_{\text{RWTAc-$\epsilon$}}$ for $\epsilon\in\{0, 0.05, 0.15, 0.25, 0.35\}$) using the same training set. For each of the policies, we then evaluate these metrics:
\begin{itemize}
	\item \textbf{MSE with respect to the the trajectories of the expert.} 	%
	For each of the trained policies, we obtain the optimal assignment  between the trajectories of the expert and the student using the cost matrix $\boldsymbol{D}_\mathbf{p}$~\cite{crouse2016implementing}. The trajectories of the student are then ranked according to the position MSE loss with respect to their assigned expert trajectory, and the index of this ranking is denoted as $\kappa$. \explanationcaserzero{}. The results are shown in Fig.~\ref{fig:comparison_estimation_hypothesis_rwta}, where 
	values above~$1$ represent cases where LSA (our approach) performs better. %
	\summaryRWTAMSEcomparison{}.
	
	\item \textbf{Number of collision-free trajectories obtained.} Using the same testing scenario as in Section~\ref{sec:comp_time_cost_deep_panther} (Fig.~\ref{fig:comparison_safe_trajs_rviz}), Fig.~\ref{fig:comparison_safe_trajs} shows the number of collision-free trajectories produced by each algorithm. Our approach is able to produce at least one collision-free trajectory for all the $\boldsymbol{g}_{\text{term}}$ tested, while RWTAr-$\epsilon$ ($\epsilon\in\{0.25, 0.35\}$) and RWTAc-$\epsilon$ ($\epsilon\in\{0, 0.05, 0.15, 0.25, 0.35\}$) fail to generate a collision-free trajectory for some of the goals, especially for the ones that are directly behind the obstacle.  
\end{itemize}

\subsection{Replanning with \addrevision{a} Dynamic Obstacle}\label{sec:replanning_dynamic_deep_panther}

We train the student in an environment that consists of a dynamic obstacle flying a trefoil-knot trajectory~\cite{trefoil2020}. The position, phase, and scale of this trefoil-knot trajectory, together with the terminal goal, are randomized. We use the Dataset-Aggregation algorithm (DAgger)~\cite{ross2011reduction} to collect the data and train the student. DAgger is an iterative dataset collection and policy training method that helps reduce covariate shift issues by querying actions of the expert while executing a partially trained policy. The total number of (observation, expert action) pairs collected is approximately $23$K.

To test this trained policy, we deploy a dynamic obstacle following a trefoil-knot trajectory with a random phase,  and manually select random $\boldsymbol{g}_{\text{term}}$. This makes the UAV replan from different initial positions, velocities, and accelerations, different states of the obstacle, and different goals. Some snapshots of the resulting collision-free trajectories generated by the student, together with the depth image of the onboard camera, are shown in Fig.~\ref{fig:snapshots_dyn_obs}. As explained in Section~\ref{sec:testing}, the collision-free trajectory that has the smallest augmented cost is the one chosen for execution.

\subsection{Generalization to other Obstacle Trajectories}\label{sec:robustness_deep_panther}

\begin{figure}
	\definecolor{mygreen_deep_panther_generalization}{RGB}{113, 207, 22}
	\definecolor{myblue_deep_panther_generalization}{RGB}{24, 252, 237}
	\begin{centering}
		\includegraphics[width=1\columnwidth]{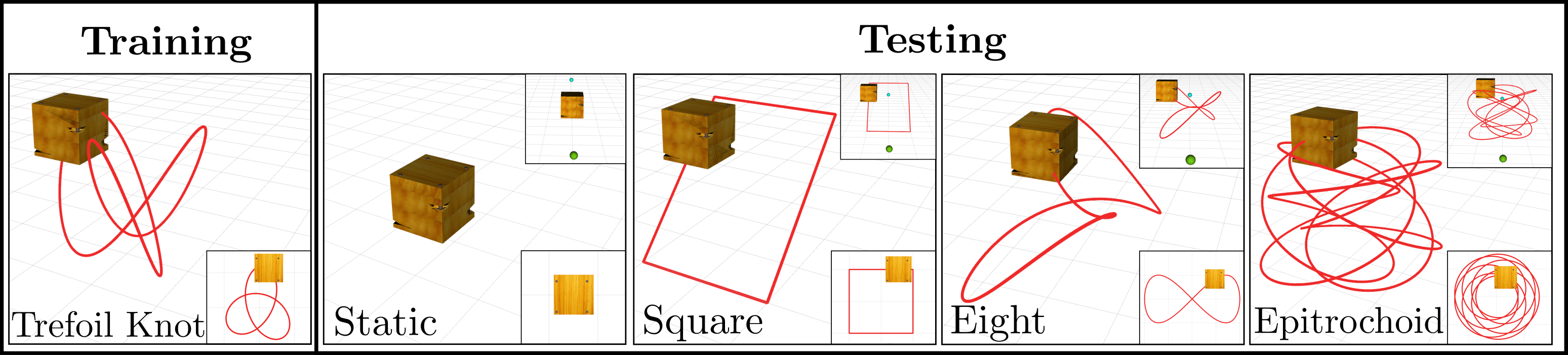}
		\par\end{centering}
	\caption[Trajectories used to train and test the student]{Trajectories used to train and test the student. 
		The task for the UAV is to fly back and forth between the two goals \tikzcircle[black,fill=mygreen_deep_panther_generalization]{25pt} and \tikzcircle[black,fill=myblue_deep_panther_generalization]{25pt}.	\label{fig:different_trajs_obstacle} } 
	\vskip-1.0ex
\end{figure}

To evaluate how well the student in Section~\ref{sec:replanning_dynamic_deep_panther} (trained using trefoil-knot obstacle trajectories) generalizes, we test it with different obstacle trajectories: static, square, eight and epitrochoid (see Fig.~\ref{fig:different_trajs_obstacle}).   During 45 seconds, the UAV must fly back and forth between two goals separated $10$~m, with the trajectory of the obstacle lying between these goals. The number of collision-free trajectories generated is shown in Table~\ref{tab:generalization_deep_panther}. Despite being trained with a different obstacle trajectory, the policy succeeded in generating at least one collision-free trajectory in all the approximately $740$ replanning steps. In all the cases the UAV reached 8 goals during the total simulation time.

\begin{table}
	\begin{centering}
		\caption{Percentage of collision-free trajectories produced by the student for different obstacle trajectories.
			The student was trained using only trefoil-knot obstacle trajectories.\label{tab:generalization_deep_panther}}
		\par\end{centering}
	\noindent\resizebox{1.0\columnwidth}{!}{%
		\begin{centering}
			\begin{tabular}{cccccc}
				\cmidrule{2-6} \cmidrule{3-6} \cmidrule{4-6} \cmidrule{5-6} \cmidrule{6-6} 
				& \textbf{Trefoil} & \textbf{Static} & \textbf{Square} & \textbf{Eight} & \textbf{Epitrochoid}\tabularnewline
				\midrule 
				$\left(n_{s}\right)_{\text{coll. free}}=0$ & \textbf{0\%} & \textbf{0\%} & \textbf{0\%} & \textbf{0\%} & \textbf{0\%}\tabularnewline
				\midrule 
				$\left(n_{s}\right)_{\text{coll. free}}\in\{1,2,3\}$ & 6\% & 0\% & 7\% & 2\% & 2\%\tabularnewline
				\midrule 
				$\left(n_{s}\right)_{\text{coll. free}}\in\{4,5,6\}$ & 94\% & 100\% & 93\% & 98\% & 98\%\tabularnewline
				\bottomrule
			\end{tabular}
			\par\end{centering}
	}
	\vskip -9.3pt
\end{table}

\subsection{\addrevision{Several Obstacles}}\label{sec:several_obstacles}

\begin{figure}
	\begin{centering}
		\addrevisiongraphics{\includegraphics[width=1\columnwidth]{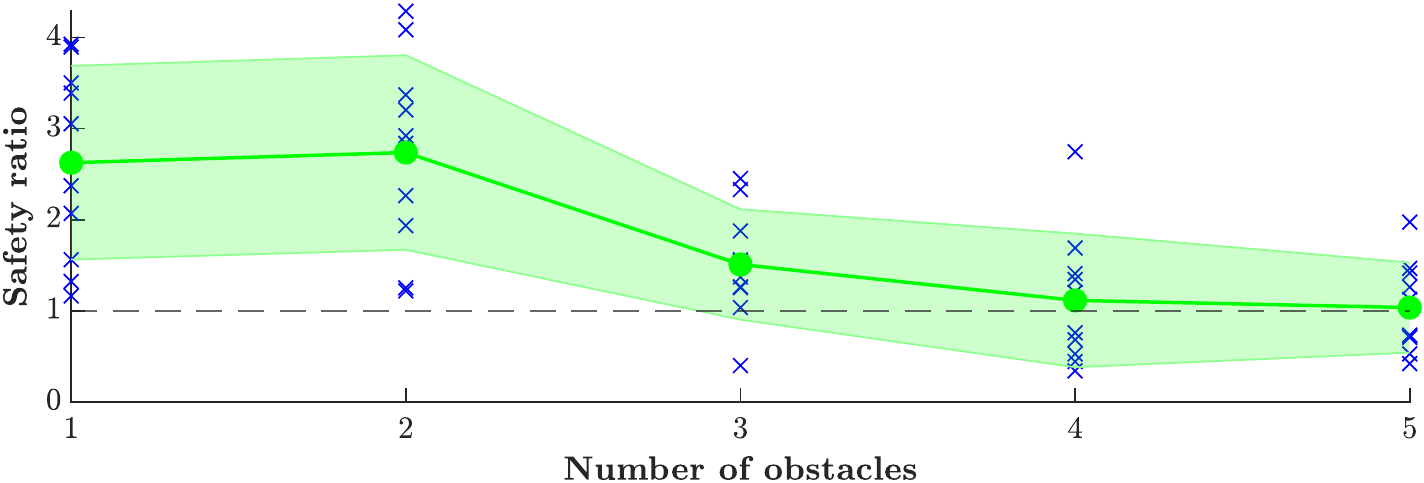}}
		\par\end{centering}
	\caption{\addrevision{Safety ratio for simulations with several obstacles. Ten random simulations were run per number of obstacles. The circles represent the mean, while the shaded area is the 1-$\sigma$ interval ($\sigma$ is the standard deviation).	Safety is ensured if the safety ratio is greater than one.} \label{fig:several_obstacles_safety_ratio} } 
	\vskip-3.0ex
\end{figure}

\addrevision{
    In these simulations, the task is to fly from $x=0$~m to $x=15$~m avoiding multiple randomly-deployed obstacles that follow epitrochoid trajectories. The policy used is the one of Section~\ref{sec:replanning_dynamic_deep_panther}, which was trained with only one obstacle that followed a trefoil-knot trajectory. \addlastRevision{For the input of the neural network, Deep-PANTHER then chooses the obstacle that has the highest probability of collision~\cite{tordesillas2021panther}}. The results, in terms of the safety ratio,\footnote{
    \addrevision{
Denoting $\boldsymbol{\kappa}_{i}(t):=\mathbf{p}(t)-\mathbf{p}_{\text{obs}\;i}(t)$
and $\boldsymbol{\rho}_{i}:=(\boldsymbol{s}_{\text{UAV}}+\boldsymbol{s}_{\text{obs}\;i})/2$,
the safety ratio is defined as 
$
\min_{t,i,j\in\{x,y,z\}}\frac{|\left(\boldsymbol{\kappa}_{i}(t)\right)_{j}|}{\left(\boldsymbol{\rho}_{i}\right)_{j}}
$. Safety is ensured if safety ratio~$>1$.}
    } are available in Fig.~\ref{fig:several_obstacles_safety_ratio}. Note how even though Deep-PANTHER has been trained with only one obstacle, it is able to succeed at all times when the number of obstacles is 1 or 2. \addlastRevision{When the number of obstacles is 3, 4, or 5, Deep-PANTHER is able to succeed on average. The failures could be addressed by incorporating multiple obstacles in the training (instead of only one obstacle), which  is left as future work.} }

\section{Conclusion and Future Work}\label{sec:conclusions}
This work derived Deep-PANTHER, a learning-based perception-aware trajectory planner in dynamic environments. Deep-PANTHER is able to achieve a similar cost as the optimization-based expert, while having a computation time two orders of magnitude faster. The multimodality of the problem is captured by the design of a loss function that assigns a distinct student trajectory to each expert trajectory. This leads to MSE losses with respect to the expert up to 18 times smaller than the (Relaxed)~Winner-Takes-All approaches. Deep-PANTHER also performs well in environments where the obstacle follows a different trajectory than the one used in training. Future work includes the extension to multiple dynamic obstacles, the inclusion of the camera images directly in the observation\addrevision{, and real-world experiments}.

\FloatBarrier

\selectlanguage{english}%
\bibliographystyle{IEEEtran}
\bibliography{my_bib}

\selectlanguage{american}%

\begin{center}
\par\end{center}

\end{document}